\documentclass[10pt,twocolumn,letterpaper]{article}

\usepackage[pagenumbers]{cvpr} 

\usepackage{multirow}
\usepackage{mathtools}
\usepackage{tabularx}
\usepackage{booktabs}
\usepackage{xcolor}
\usepackage{makecell}
\usepackage{bbm}
\usepackage{ulem}  
\usepackage{stfloats}
\usepackage{placeins}
\usepackage{colortbl}
\usepackage[table]{xcolor}
\definecolor{cvprblue}{rgb}{0.21,0.49,0.74}
\usepackage[pagebackref,breaklinks,colorlinks,allcolors=cvprblue]{hyperref}
\usepackage{tabularx}

\def\paperID{*****} 
\def\confName{CVPR}
\def\confYear{2026}

\title{SIGMMA: Hierarchical Graph-Based Multi-Scale Multi-modal Contrastive Alignment of Histopathology Image and Spatial Transcriptome}

\author{
\parbox{\textwidth}{
\centering
Dabin Jeong$^{1}$,
Amirhossein Vahidi$^{1,2}$,
Ciro Ramírez-Suástegui$^{1}$,
Marie Moullet$^{1,2}$,
Kevin Ly$^{1,2}$,\\
Mohammad Vali Sanian$^{1}$,
Sebastian Birk$^{1,2,3}$,
Yinshui Chang$^{1}$,
Adam Boxall$^{1}$,
Daniyal Jafree$^{1}$,\\
Lloyd Steele$^{1}$,
Vijaya Baskar MS$^{1}$,
Muzlifah Haniffa$^{1}$\footnotemark[1],
Mohammad Lotfollahi$^{1,2,4}$\footnotemark[1] \\[10pt]
Wellcome Sanger Institute, Wellcome Genome Campus, Cambridge, UK$^{1}$ \\
Cambridge Centre for AI in Medicine, University of Cambridge, Cambridge, UK$^{2}$ \\
Institute of AI for Health, Helmholtz Center Munich, Neuherberg, Germany$^{3}$ \\
Cambridge Stem Cell Institute, University of Cambridge, Cambridge, UK$^{4}$ \\
{\tt\small \{mh32,ml19\}@sanger.ac.uk}
}}

\begin{document}
\maketitle
\renewcommand{\thefootnote}{\fnsymbol{footnote}}
\footnotetext[1]{Corresponding authors}
\begin{abstract}
Recent advances in computational pathology have leveraged vision–language models to learn joint representations of Hematoxylin and Eosin (HE) images with spatial transcriptomic (ST) profiles.
However, existing approaches typically align HE tiles with their corresponding ST profiles at a single scale, overlooking fine-grained cellular structures and their spatial organization.
To address this, we propose \textsc{Sigmma}, a multi-modal contrastive alignment framework for learning hierarchical representations of HE images and spatial transcriptome profiles across multiple scales. \textsc{Sigmma} introduces multi-scale contrastive alignment, ensuring that representations learned at different scales remain coherent across modalities. Furthermore, by representing cell interactions as a graph and integrating inter- and intra-subgraph relationships, our approach effectively captures cell–cell interactions, ranging from fine to coarse, within the tissue microenvironment. We demonstrate that \textsc{Sigmma} learns representations that better capture cross-modal correspondences, leading to an improvement of avg. 9.78\% in the gene-expression prediction task and avg. 26.93\% in the cross-modal retrieval task across datasets. We further show that it learns meaningful multi-tissue organization in downstream analyses. 
\end{abstract}


\section{Introduction}
\label{sec:intro}


\begin{figure}
  \centering
    \includegraphics[width=1\linewidth]{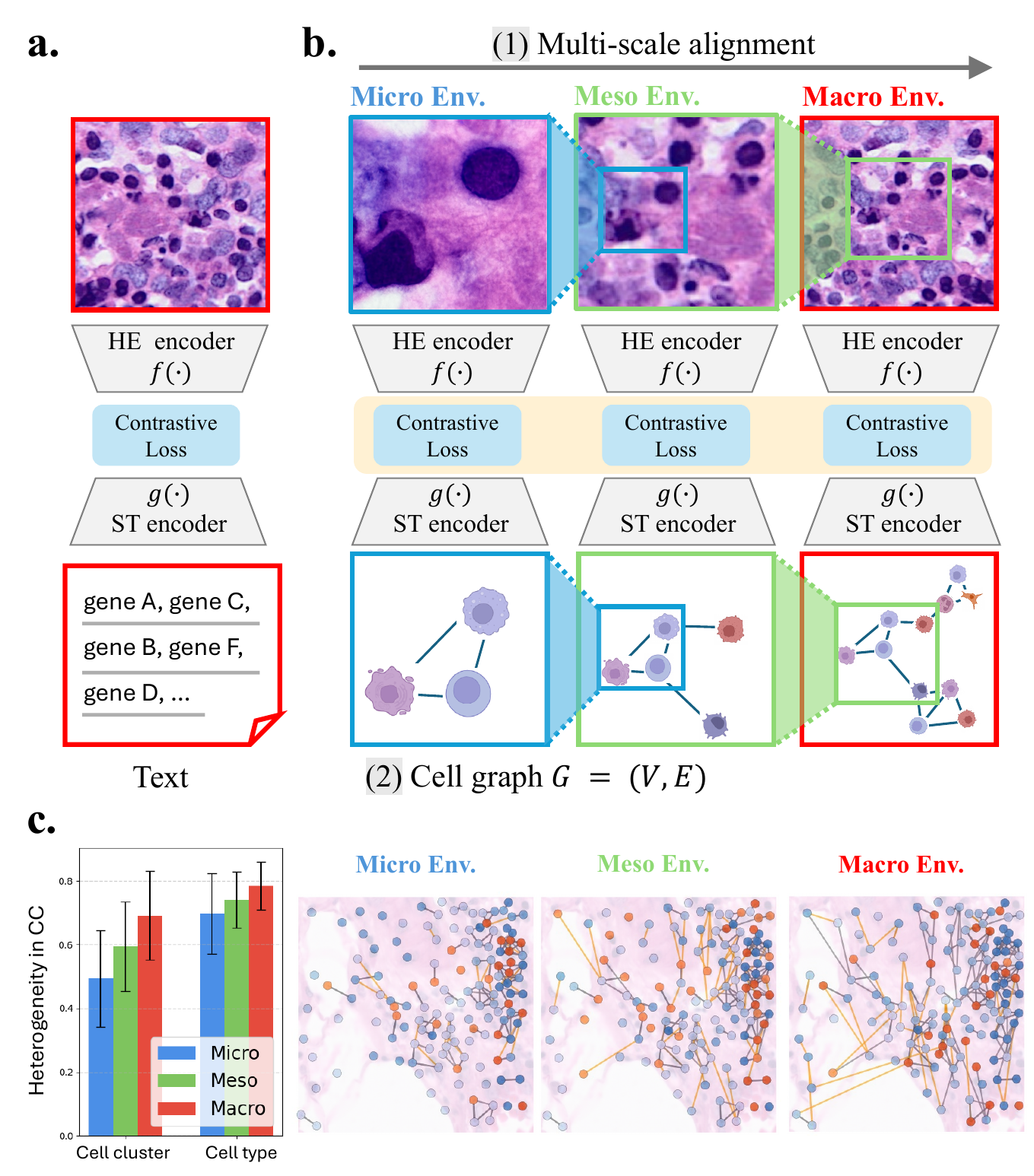}
    \vspace{-20pt}
    \caption{\textbf{Motivation.} \textbf{(a)} Limitations of previous vision-language model-based HE-ST alignment. \textbf{(b)} How \textsc{Sigmma} addresses these limitations by (1) multi-scale alignment and (2) adopting a cell graph structure that preserves 2D coordinates and cell-cell relationships. \textbf{(c)} {\textsc{Sigmma} captures multi-scale information, with ST representations of each cell becoming more heterogeneous at larger scales.} CC,connected component.}
    \vspace{-5pt}
    \label{fig:motivation}
\end{figure}



Tissue architecture is hierarchically organized across multiple spatial scales, from the \textit{micro environment} (Fig. \ref{fig:motivation} blue) comprising small clusters of interacting cells within local regions, to the \textit{meso environment} (Fig. \ref{fig:motivation} green) encompassing cellular neighborhoods of dozens of cells, and up to the \textit{macro environment} (Fig. \ref{fig:motivation} red) characterized by macroscopic structures, e.g., tertiary lymphoid structures \citep{teillaud2024tertiary}. Understanding these hierarchical contexts requires both morphological and molecular views of tissue.

Computational histopathology has been advanced by large-scale analysis of cellular morphology, tissue composition, and spatial organization, most commonly Hematoxylin Eosin (HE) images. In parallel, single-cell spatial transcriptomics (ST) enables the molecular profiling of individual cells with 2D spatial coordinates, providing direct links between morphology and gene expression profiles. Together, these technologies allow researchers to map how individual cell types, cell-cell interactions, and spatial organization of cells contribute to physiological and pathological processes \citep{zhang2025thor}. {Multi-modal learning of HE images with ST profiles provides a unified representation of tissue morphology and molecular state, enabling the identification of molecular heterogeneity that is not discernible from HE images alone (Fig. \ref{fig:motivation}c).}

\paragraph{Why do we need a graph structure for ST modeling?} Recent contrastive learning approaches learn joint representations of HE images and ST in a shared space to support cross-modal tasks, including image-to-expression retrieval and image-to-expression prediction. Notably, vision–language (VL) models originally developed for HE–biomedical text (e.g., caption, scientific papers, clinical notes) have been adapted for ST \citep{chen2025visual, glettig2025h, zou2025predicting}. However, current VL-based HE–ST contrastive alignment approaches represent ST as a 1D gene sequence aggregated across cells \citep{chen2025visual, glettig2025h, zou2025predicting}, thereby inevitably discarding the original 2D spatial organization and cell–cell interactions within the tissue (Fig. \ref{fig:motivation}a, bottom). In contrast, graph representations inherently encode spatial topology and relational structure, enabling explicit modeling of cell–cell interactions and the surrounding {tissue context} (Fig. \ref{fig:motivation}b bottom).

\paragraph*{\textbf{Why do we need hierarchical multi-scale HE-ST contrastive alignment?}}
The hierarchical organization of tissue makes multi-scale HE–ST alignment inherently challenging. Multi-scale alignment implicitly requires correspondence across region-of-interest (ROI) granularities. Embeddings learned from contrastive loss (e.g., InfoNCE loss) maximize the lower bound of mutual information between two pairs \citep{oord2018representation}, which tends to emphasize salient features at the expense of finer details \citep{yang2022vision}. Specifically, graph-structured ST data complicates multi-scale alignment because message passing expands the receptive field based on graph connectivity rather than image ROI scales, leading to mismatched spatial scopes across modalities.
To overcome this limitation, multi-scale contrast alignment methods incorporate multiple ROI sizes, thus capturing both coarse- and fine-grained tissue features.  (Fig. \ref{fig:motivation}b).

\paragraph*{\textbf{Our contribution.}}
We propose \textsc{Sigmma}, a hierarchical, graph-based, multi-scale alignment framework for HE–ST.

\begin{itemize}
\item \textbf{Graph-structured representation of ST.} We represent ST as a cell graph that preserves spatial topology and the structure of cell-cell relationships. A hierarchical graph module integrates intra- and inter-subgraph relationships to capture local neighborhoods and long-range dependencies that are lost in sequence-based ST representations.
\item \textbf{Multi-scale cross-modal alignment.} We introduce a multi-scale HE–ST alignment framework that enforces alignment consistency across multiple spatial resolutions. Our {multi-scale contrastive} objective aligns representations from micro, meso, and macro contexts, improving fine-grained and coarse-grained correspondence.
\item \textbf{ST graph–HE image scale reconciliation.} {We progressively expand the graph receptive field through hierarchical graph learning, matching it to the image ROI size and enabling consistent correspondence between modalities across scales.}
\item \textbf{Performance improvements and interpretability.} \textsc{Sigmma} yields improvements in downstream tasks, including gene-expression prediction and image–expression retrieval,  across five datasets and produces embeddings that reveal biologically meaningful tissue organization.
\end{itemize}

\section{Related Work}

\paragraph*{ST at single-cell resolution.}
ST has emerged as a powerful approach to map gene expression within the spatial context of tissues. Specifically, it measures gene expression together with 2D spatial coordinates, indicating the location and level of expression of specific genes. There are two main techniques for measuring ST: Visium \citep{williams2022introduction} and Xenium \citep{janesick2023high}. Visium is a sequencing-based platform which captures transcriptomic signals at the spot level, where each spot typically aggregates the expression profiles of multiple neighboring cells. 
In contrast, the Xenium platform utilizes high-resolution \textit{in situ} hybridization and imaging to measure gene expression at the cellular/subcellular levels, offering deeper insights into cell–cell interactions.

In this work, we use Xenium rather than Visium because Xenium provides cell-level spatial transcriptomics, enabling alignment with HE images while explicitly modeling each cell’s 2D spatial context.

\paragraph*{Tiling of HE WSI image.}
Whole-slide images (WSIs) are gigapixel-scale, making direct application of vision models computationally prohibitive and forcing heavy down-sampling that removes critical cellular-level signals \citep{hou2016patch}. Since discriminative patterns are small, sparse, and spatially scattered, tile-level modeling enables vision models to learn high-resolution local features by training on small image tiles, and leads to WSI-level tasks by aggregating tile-level embeddings \citep{azizi2023robust, hoptimus0, chen2024towards, jaume2024transcriptomics, jaume2024hest}. 
As molecular phenotypes and cellular contexts vary across localized regions, tile-level alignment can provide a more fine-grained correspondence between image features and transcriptomic signals than slide-level alignment. 

Motivated by this, our work focuses on tile-level alignment between HE and ST features, enabling cross-modal learning at a spatially-resolved and fine-grained level.

\paragraph*{Foundation models for HE and ST.}
Foundation models have recently emerged in computational histopathology for both HE images and ST.
For HE image, DINO\citep{oquab2023dinov2}-based vision foundation models enable scalable learning of morphology-rich representations that generalize across slides \citep{chen2024towards, hoptimus0, vorontsov2024foundation}, .
Extending this line of work, hierarchical transformers leverage the intrinsic multi-scale structure of WSIs and learn representations across cellular, tissue, and slide levels \citep{chen2022scaling}. 
In parallel, ST foundation models, inspired by large language models, learn cell-level representations by treating gene expression profiles as sequences using transformer architectures \citep{wang2025scgpt, birk2025quantitative, seb2025STEMO, blampeyNovae2024}.

These uni-modal foundation models provide generalizable representations for HE and ST, serving as building blocks for downstream multi-modal alignment. In this work, we build upon these foundation models to learn a unified cross-modal representation between HE and ST.

\paragraph{HE-ST contrastive alignment.}
Early attempts to predict ST profile directly regressed spot-level expression from HE image using convolutional neural networks or transformer backbones \citep{chung2024accurate, ganguly2025merge}. Recent methods introduced spatial graphs, representing spots as nodes connected by proximity and formulated the ST prediction problem as node-level regression task \citep{ganguly2025merge}. With the advent of high-resolution ST, the paradigm has shifted from spot to cellular/subcellular-level modeling, leading to cell-graph approach \citep{ge2025deep} and diffusion-based image-to-expression generation at subcellular resolution \citep{xu2025topocellgen}. In parallel, contrastive learning–based approaches have emerged that align HE and ST modalities rather than predicting one from the other, enriching cross-modal representations and improving downstream prediction \citep{xie2023spatially, redekop2025spade}. VL frameworks extend contrastive alignment, pairing HE tiles with biomedical text or gene-token sequences to learn joint representations \citep{huang2023visual, lu2024visual, albastaki2025multi}.
Recent works leverage ST to perform spatially resolved alignment between image regions and Visium spot-level expression \citep{chen2025visual, gindra2025large, guojade}, with subsequent studies extending this to cell-level alignment with Xenium data \citep{glettig2025h}.

In contrast, our framework introduces graph-based multi-scale alignment between HE and ST. We represent each ST tile as a cell graph constructed from cell coordinates and perform alignment with HE tile at multiple spatial scales, maintaining spatial consistency and enabling fine-grained cell-level correspondence across modalities.

\begin{figure*}[t]
  \centering
  \includegraphics[width=0.99\linewidth]{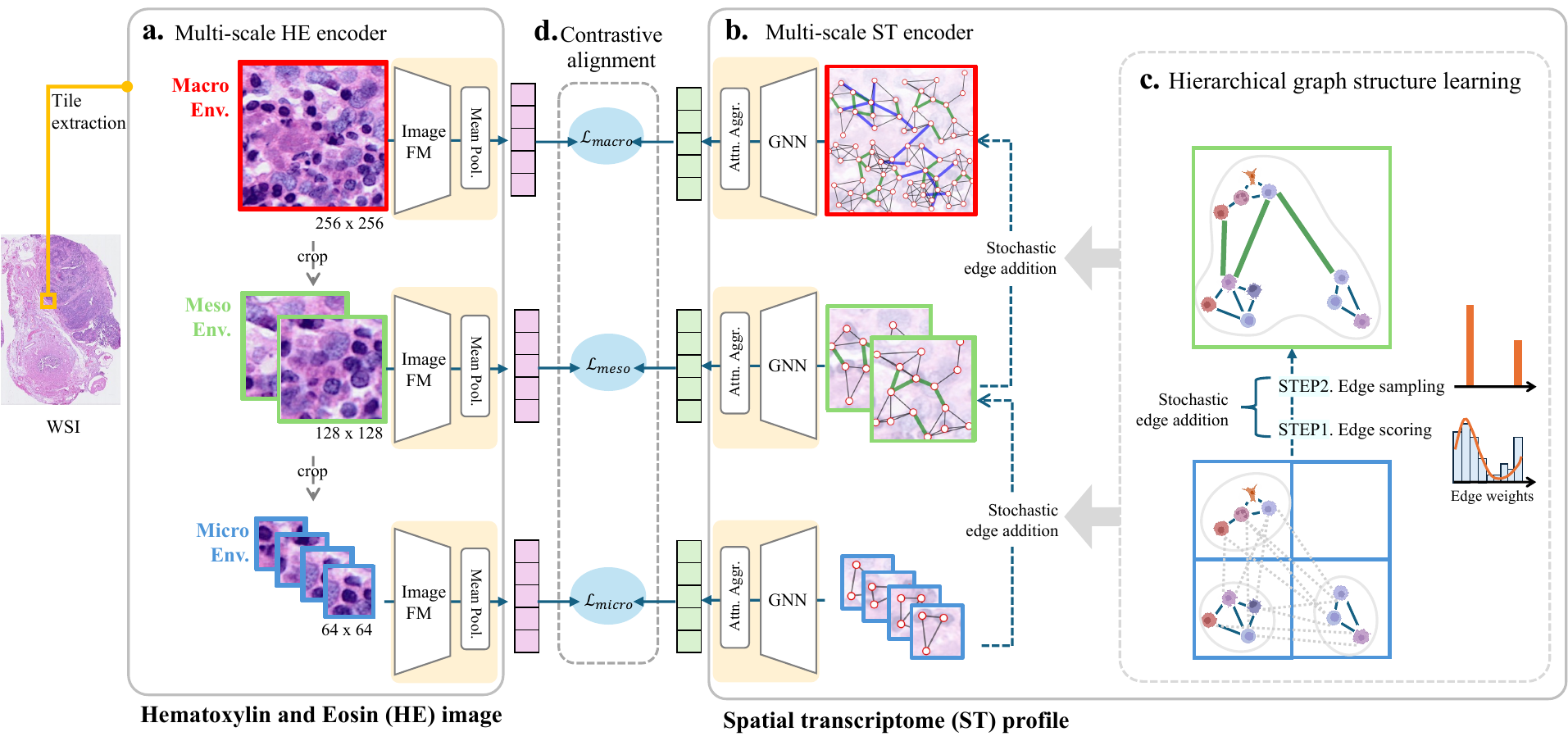}
  \vspace{-10pt}
  \caption{\textbf{Schematic overview of \textsc{Sigmma}}. Given a tessellated tile of HE and ST, \textsc{Sigmma} aligns HE-ST tiles at multi-scale. \textbf{(a)} For HE side, multi-crop strategy is applied (Sec. \ref{subsection:HE_encoder}). \textbf{(b)} hierarchical graph structure learning is applied for ST side (Sec. \ref{subsection:ST_encoder}). \textbf{(c)} Hierarchical graph structure learning consists of stochastic edge addition with a neighbor-patch constraint. \textbf{(d)} Multi-modal multi-scale contrastive alignment of HE and ST (Sec. \ref{subsection:multimodal-alignment}). FM,foundation model;GNN,graph neural network.} 
  \label{fig:schematic_overview}
\end{figure*}

\section{Problem Definition}
We consider paired HE images and ST profiles obtained from the same tissue section $k$, denoted by $(\mathcal{I}^k, \mathcal{S}^k)$, where each spans $H_k \times W_k$ pixels, with $H_k$ and $W_k$ representing height and width of the section. For simplicity, we omit the section index $k$. 
$\mathcal{I}$ and $\mathcal{S}$ are tessellated into $m \times m$ pixel sized tiles, $\{(I_i, S_i)\}_{i=1}^{n}$, where $n = \left\lfloor \frac{H}{m} \right\rfloor \times \left\lfloor \frac{W}{m} \right\rfloor$ is the number of tiles extracted in a WSI. We train an HE image encoder $f(\cdot)$ and an ST encoder $g(\cdot)$, each parameterized by a neural network, where the encoders yield latent HE embedding $z_i^I = f(I_i)$ and latent ST embedding $z_i^S = g(S_i)$,  respectively. The objective is to jointly optimize the HE image and ST encoders such that paired HE and ST embeddings, $(z_i^I, z_i^S)$, are aligned in a shared latent space, thereby capturing the cross-modal correspondence between HE and ST.


\section{Preliminaries}
\paragraph*{Graph neural network.}
Let $G = (V, E)$ be a graph where $V$ denotes the set of nodes and $E$ the set of edges.
Graph neural networks (GNN) learn node representations through iterative message passing between connected nodes.
We denote the embedding of node $v \in V$ at layer $l$ as $h_v^{(l)}$, and define $\mathcal{N}(v)$ as the set of neighboring nodes of $v$ determined by the edge set $E$.
At the $l$-th iteration, each node updates its embedding by aggregating information from $\mathcal{N}(v)$ as follows \citep{hamilton2017inductive}:
\begin{align*}
h_{\mathcal{N}(v)}^{(l)} &= \text{Aggregate}\left(\{\, h_u^{(l)} , \forall u \in \mathcal{N}(v) \,\}\right) \\
h_v^{(l+1)} &= \sigma\!\left( W^{(l)} \cdot \text{Concat}\!\left(h_v^{(l)}, h_{\mathcal{N}(v)}^{(l)}\right) \right)
\end{align*}
where $W$ denotes a learnable weight matrix and $\sigma$ denotes a non-linear activation function. 

\paragraph*{Stochastic edge addition for GNN.}
\label{subsubection: stochastic_edge_selection}
Stochastic edge addition enables adaptive graph sparsification and has been applied to document graphs \citep{piao2022sparse} and chemical graphs \citep{piao2024improving}. 
Given node embeddings $h_u$ and $h_v$ learned via GNN, a probability distribution function $\phi
(\cdot)$ for edge selection is defined as follows:
\begin{equation*}
    s_{uv} = \phi(h_u, h_v) = \sigma\big(\text{MLP}([h_u, h_v])\big)
\end{equation*}
where $[ \cdot, \cdot]$ denotes the concatenation operator, $\sigma$ denotes a non-linear activation function. 
Here, $s_{uv}$ is the score indicating the likelihood of forming an edge between nodes $u$ and $v$.

To create a stochastic edge selector from the score, a binary variable $p_{uv} \in \{0, 1\}$ is drawn from the Bernoulli distribution $p_{uv} \sim \{\ \pi_1 \coloneq s_{uv}, \pi_0 \coloneq 1 - s_{uv} \}$.
The Gumbel softmax relaxation \citep{jang2016categorical} is applied to make edge selection differentiable. The differentiable edge selection probability is thus defined as follows:
\begin{equation*}
    \hat{p}_{uv} = \frac{\exp((\log \pi_1+g_1)/\tau)}{\sum_{i \in \{0,1\}} \exp((\log \pi_i+g_i)/\tau)}
\end{equation*}
where $g_1$ and $g_0$ are i.i.d. variables sampled from the Gumbel distribution, and $\tau$ denotes the temperature hyperparameter that controls the spikiness of the relaxed Bernoulli distribution.

\section{Proposed Method: \textbf{\textsc{Sigmma}}}
Here, we present \textsc{Sigmma}, a framework for Spatial transcriptome–histology Image representation learning via hierarchical Graph-based Multi-scale Multi-modal Alignment (Fig. \ref{fig:schematic_overview}). \textsc{Sigmma} consists of three components: a multi-scale HE encoder (Sec. \ref{subsection:HE_encoder}), a multi-scale ST encoder (Sec. \ref{subsection:ST_encoder}), and a multi-modal multi-scale contrastive alignment component (Sec. \ref{subsection:multimodal-alignment}).


\subsection{Multi-scale HE encoder}
\label{subsection:HE_encoder}
To capture the hierarchical spatial contexts of an HE tile, which is an RGB image $I_i \in \mathbb{R}^{m \times m \times 3}$, we adopt a multi-crop strategy \citep{guo2024llava, liu2024improved} using pretrained image encoders $f(\cdot)$ (Fig. \ref{fig:schematic_overview}a). For simplicity, we omit the tile index $i$ in this section and the following section.
Each HE image tile $I$ is partitioned into a varying grid size that captures various ROI scales (Fig. \ref{fig:schematic_overview} red, green, blue box): 
$4 \times 4$ grid micro patches $\{I_{\text{micro},j}\}_{j=1}^{16}$
, $2 \times 2$ grid meso patches $\{I_{\text{meso},j}\}_{j=1}^4$
, and a macro patch $\{I_{\text{macro},j}\}_{j=1}^{1}$, where $j$ enumerates patches within each grid in this section.

After extracting patch features from each ROI scale, we resize and interpolate each patch to a unified scale, matching the training resolution of the HE foundation model \citep{chen2024towards}. At each micro, meso, and macro scale, the resulting patch embeddings are then mean-pooled to obtain tile-level image embeddings, $z_{\text{micro}}, z_{\text{meso}}, z_{\text{macro}}$, which capture local to broader spatial contexts, respectively.
\begin{gather*}
z_{\text{micro}}^I = \text{Pool}_I(f({I_{\text{micro}}})) \\
z_{\text{meso}}^I = \text{Pool}_I(f({I_{\text{meso}}})) \\
z_{\text{macro}}^I = \text{Pool}_I(f({I_{\text{macro}}}))
\end{gather*}
where $\text{Pool}_I(\cdot)$ is a grid-wise mean pooling operator. 

\subsection{Multi-scale ST encoder}
\label{subsection:ST_encoder}
\paragraph{Graph Representation of ST.}
Given an $m \times m$ pixel-sized ST tile, $S$, we can detect 2D coordinates of individual cells within the tile. Then, $S$ can be represented as a cell graph $G = (V, E)$, where $V$ denotes cells, and $E$ denotes edges that capture cell-cell interactions. Node embeddings are initialized by the ST foundation model \citep{seb2025STEMO}.

\paragraph{Hierarchical graph structure learning.}
We adapted a stochastic edge addition algorithm (Sec. \ref{subsubection: stochastic_edge_selection}) to reconcile the difference in granularity between HE and ST. To this end, we hierarchically expand a small subgraph by linking its neighbors (Fig. \ref{fig:schematic_overview}c).

Specifically, we first extract a subgraph $G_{\text{micro}} = (V_{\text{micro}}, E_{\text{micro}})$, where $V_{\text{micro}} \subseteq V$ and $E_{\text{micro}} \subseteq E$, such that cells within the corresponding image tile $I_{\text{micro}}$ are connected by edges defined based on spatial proximity \citep{squidpy_spatial_neighbors}. Edges are stochastically added by a stochastic edge-addition layer $\psi^{(l)}(\cdot)$ followed by GNN layers. In each layer $\psi^{(l)}(\cdot)$, given each node $u$, neighbor nodes are updated by sampling from the candidate set, and edges are connected to the selected nodes as follows:
\begin{gather*}
    \mathcal{N}_{\text{meso}}(u)^{(l-1)} = \mathcal{N}_{\text{micro}}^{(l-2)}(u) \cup \{\, v \mid \forall v \rightarrow p_{uv}^{(l-1)} = 1 \,\} \\
    \mathcal{N}_{\text{macro}}(u)^{(l)} = \mathcal{N}_{\text{meso}}^{(l-1)}(u) \cup \{\, v \mid \forall v \rightarrow p_{uv}^{(l)} = 1 \,\}
\end{gather*}
where $v \in \mathcal{N}_{*}(v)$ denotes a neighbor of node $v$ connected through edges defined at each scale, and $p_{uv}$ denotes edge selection probability (Sec. \ref{subsubection: stochastic_edge_selection}).
{Here, instead of treating all nodes as candidates, we enforce a neighbor-patch constraint that allows edges to form only between nodes in adjacent patches.}
This yields distinct graph topologies at each scale: $G_{\text{micro}}, G_{\text{meso}}, G_{\text{macro}}$.

These steps above describe how the ST encoder $g(\cdot)$ learns node embeddings from each scale-specific graph (Fig. \ref{fig:schematic_overview}b).
Given the node embeddings learned from graph topology, we obtain graph-level representations as follows,
\begin{gather*}
z_{\text{micro}}^S = \text{Pool}_S(g({G_{\text{micro}}})) \\
z_{\text{meso}}^S = \text{Pool}_S(g({G_{\text{meso}}})) \\
z_{\text{macro}}^S = \text{Pool}_S(g({G_{\text{macro}}}))
\end{gather*}
where $\text{Pool}_S(\cdot)$ is global attention pooling operator \citep{li2015gated} over nodes.

\paragraph{{Neighbor-patch constrained edge addition.}}
\label{paragraph:neighbor-constraint}
Here, we elaborate on how spatial constraints restrict edge addition to neighboring subgraphs.
Given cell coordinates $(x_p, y_q)$ on a 2D grid, we divide the grid into local blocks of size $b \times b$.
Each node $p$ belongs to a block indexed as follows:
\begin{equation*}
    b_x(p) = \left\lfloor \frac{x_p}{b} \right\rfloor, 
    \qquad
    b_y(q) = \left\lfloor \frac{y_q}{b} \right\rfloor.
\end{equation*}
An edge $(p, q)$ is allowed only if both nodes lie within the same block, i.e.,
\begin{equation*}
    \label{eq:intra_block}
    \mathbbm{1}_{\text{intra}}(p, q) =
    \begin{cases}
    1, & \text{if } b_x(p) = b_x(q) \text{ and } b_y(p) = b_y(q), \\
    0, & \text{otherwise.}
    \end{cases}
\end{equation*}
This constraint enforces edge connectivity only within each $b \times b$ local grid, preventing cross-block edges. $b{=}1,2,4$ for macro, meso, and micro scale, respectively.

\subsection{Multi-modal multi-scale contrastive alignment}
\label{subsection:multimodal-alignment}
Contrastive learning, a mainstream of self-supervised learning, has been extended to multi-modal domains \citep{radford2021learning, lu2024visual}.
In our framework, the objective of contrastive learning is to train the two encoders $f(\cdot)$ and $g(\cdot)$ jointly that maximizes alignment of the latent representations of paired HE and ST tiles $(I_i, S_i)$ while minimizing similarity across unmatched pairs. 
We utilized InfoNCE loss \citep{oord2018representation} to achieve this objective:
\begin{gather*}
\mathcal{L}_{I\to S}
= - \frac{1}{N} \sum_{i=1}^N \log 
\frac{
    \exp\left(\mathrm{sim}(z_i^I, z_i^S) / \tau\right)
}{
    \sum_{j} \exp\left(\mathrm{sim}(z_i^I, z_j^S) / \tau\right)
} \\[8pt]
\mathcal{L}_{S\to I}
= - \frac{1}{N} \sum_{i=1}^N \log 
\frac{
    \exp\left(\mathrm{sim}(z_i^S, z_i^I) / \tau\right)
}{
    \sum_{j} \exp\left(\mathrm{sim}(z_i^S, z_j^I) / \tau\right)
} \\[8pt]
\mathcal{L}_{\textsc{align}}(z^I, z^S)
= \frac{1}{2}[\mathcal{L}_{I\to S} + \mathcal{L}_{S\to I}]
\end{gather*}
where $N$ is the number of {samples} within a batch, the index $j$ runs over all samples in the batch, $\mathrm{sim}(\cdot,\cdot)$ denotes the cosine similarity between embeddings, $\tau$ is a temperature parameter controlling the sharpness of the similarity distribution.

At each scale, we compute a contrastive loss between the HE and ST tile embeddings (Fig. \ref{fig:schematic_overview}d). The micro-level loss $\mathcal{L}_{\textsc{micro}}$ is computed between the micro-scale embeddings, i.e., $\mathcal{L}_{\textsc{micro}} = \mathcal{L}_{\textsc{align}}(z_{\text{micro}}^I, z_{\text{micro}}^S)$, Similarly, the meso-level loss $\mathcal{L}_{\textsc{meso}}$ aligns the meso-scale embeddings $(z_{\text{meso}}^I, z_{\text{meso}}^S)$,
and the macro-level loss $\mathcal{L}_{\textsc{macro}}$ aligns the macro-scale embeddings, $(z_{\text{macro}}^I, z_{\text{macro}}^S)$, respectively.
The total objective function is as follows:
\begin{gather*}
    \mathcal{L} = \mathcal{L}_{\textsc{micro}} + \mathcal{L}_{\textsc{meso}} + \mathcal{L}_{\textsc{macro}}
\end{gather*}

\section{Experiments}
\label{experiments}

\paragraph*{Datasets.}
We conduct extensive benchmarking on the HEST-1k dataset \citep{jaume2024hest}, the largest publicly available dataset of paired HE and ST data. Four subsets of the dataset provide paired HE–Xenium ST data, covering four cancer types: Invasive Ductal Carcinoma (IDC), Pancreatic Adenocarcinoma (PAAD), Skin Cutaneous Melanoma (SKCM), and Lung Adenocarcinoma (LUAD). In addition to the public datasets, we include an in-house skin dataset.
Following the tiling scheme in multiple histopathology image foundation models \citep{chen2024towards, vorontsov2024foundation, hoptimus0}, we tessellate each WSI into $256 \times 256$ pixel-sized tiles at 20x magnification level, which corresponds to $ 0.5 \mu\text{m/pixel}$ resolution.  
For more details on data preprocessing/data splits, see Suppl. Sec. \ref{suppl:data}.

\paragraph*{Baselines and Evaluation metrics.}
We comprehensively compare \textsc{sigmma} against three categories of baselines:
(1) Uni-modal HE image encoder, UNI \citep{chen2024towards};
(2) Multi-modal vision–language (VL) models pre-trained on natural image–caption pairs, CLIP \citep{radford2021learning}, or medical text, PLIP \citep{huang2023visual};
(3) HE–ST contrastive alignment models, including OmiCLIP \citep{chen2025visual}, which uses a text encoder by representing ST as a 1D sequence of gene names, and BLEEP \citep{xie2023spatially}, a simple MLP operating on batch corrected ST principal components.
To ensure fair comparison, all baselines are fine-tuned on the datasets used in this study. For each method, we adopt the hyperparameters reported in the original paper; when unavailable, we determine them through grid search. For more details on the experiment setting, see Suppl. Sec. \ref{suppl:method}. 

\begin{table*}[t]
    \centering
    \scriptsize
    \setlength{\tabcolsep}{4pt} 
    \caption{Gene expression prediction performance across HEST1k and in-house datasets.}
    \vspace{-2mm}
    \begin{tabularx}{\textwidth}{c|cc|cc|cc|cc|cc}
    \toprule[1.2pt]
     Dataset 
      & \multicolumn{2}{c|}{\textit{HEST1k-LUAD}} 
      & \multicolumn{2}{c|}{\textit{HEST1k-PAAD}} 
      & \multicolumn{2}{c|}{\textit{HEST1k-SKCM}} 
      & \multicolumn{2}{c|}{\textit{HEST1k-IDC}} 
      & \multicolumn{2}{c}{\textit{in-house skin}} \\
    \cmidrule(lr){2-11}
     Model 
      & MSE (↓) & PCC (↑)
      & MSE (↓) & PCC (↑)
      & MSE (↓) & PCC (↑)
      & MSE (↓) & PCC (↑)
      & MSE (↓) & PCC (↑) \\
    \midrule[1.2pt]
      UNI & 0.046$\pm$0.041 & 0.476$\pm$0.064 & \underline{0.008$\pm$0.008} & 0.470$\pm$0.064 & 0.073$\pm$0.080 & \underline{0.666$\pm$0.032} & \underline{0.046$\pm$0.041} & \underline{0.476$\pm$0.064} & 0.094$\pm$0.072 & \underline{0.418$\pm$0.014}\\
    \cmidrule(lr){1-11}
      CLIP     & 0.052$\pm$0.052 & 0.467$\pm$0.088 & 0.009$\pm$0.010 & 0.245$\pm$0.081 & 0.080$\pm$0.066 & 0.541$\pm$0.018 & 0.052$\pm$0.052 & 0.467$\pm$0.088 & 0.103$\pm$0.084 & 0.330$\pm$0.022 \\
      PLIP     & 0.027$\pm$0.016 & 0.561$\pm$0.059 & 0.011$\pm$0.012 & 0.432$\pm$0.032 & 0.060$\pm$0.055 & 0.612$\pm$0.058 & 0.053$\pm$0.050 & 0.465$\pm$0.089 & 0.107$\pm$0.084 & 0.331$\pm$0.015  \\
    \cmidrule(lr){1-11}
      BLEEP & \textbf{0.011$\pm$0.011} & 0.252$\pm$0.082 & \textbf{0.004$\pm$0.008} & 0.124$\pm$0.137 & \textbf{0.012$\pm$0.006} & 0.594$\pm$0.232 & \textbf{0.004$\pm$0.003} & 0.443$\pm$0.159 & \textbf{0.035$\pm$0.008} & 0.292$\pm$0.034 \\
      OmiCLIP & 0.022$\pm$0.013 & \underline{0.613$\pm$0.034} & 0.018$\pm$0.016 & \underline{0.480$\pm$0.026} & 0.083$\pm$0.057 & 0.481$\pm$0.061 & 0.053$\pm$0.044 & 0.472$\pm$0.055 & 0.118$\pm$0.093 & 0.230$\pm$0.025 \\
    \cmidrule(lr){1-11}
      \rowcolor{yellow!20} \textsc{Sigmma}   & \underline{0.015$\pm$0.007} & \textbf{0.741$\pm$0.023} & 0.015$\pm$0.015 & \textbf{0.485$\pm$0.036} & \underline{0.051$\pm$0.048} & \textbf{0.744$\pm$0.052} & 0.051$\pm$0.043 & \textbf{0.510$\pm$0.072} & \underline{0.060$\pm$0.032} & \textbf{0.452$\pm$0.025} \\
    \bottomrule[1.2pt]
    \end{tabularx}
    \vspace{-3pt}
    \label{table:gex_prediction}
\end{table*}
\begin{table*}[!t]
    \centering
    \scriptsize
    \setlength{\tabcolsep}{4pt}
    \caption{Cross-modal retrieval performance across HEST1k and in-house datasets. R,recall.}
    \vspace{-2mm}
    \begin{tabular}{c|ccc|ccc|ccc|ccc|ccc}
        \toprule[1.2pt]
        & \multicolumn{15}{c}{HE $\rightarrow$ ST} \\
        \cmidrule(lr){2-16}
        Dataset 
        & \multicolumn{3}{c|}{\textit{HEST1k-LUAD}} 
        & \multicolumn{3}{c|}{\textit{HEST1k-PAAD}} 
        & \multicolumn{3}{c|}{\textit{HEST1k-SKCM}} 
        & \multicolumn{3}{c|}{\textit{HEST1k-IDC}} 
        & \multicolumn{3}{c}{\textit{in-house skin}} \\
        \cmidrule(lr){2-16}
        Model 
        & R@5\% & R@10\% & R@15\% 
        & R@5\% & R@10\% & R@15\% 
        & R@5\% & R@10\% & R@15\% 
        & R@5\% & R@10\% & R@15\% 
        & R@5\% & R@10\% & R@15\% \\
        \midrule[1.2pt]
        CLIP    & 0.278 & 0.452 & 0.566 & \underline{0.195} & \underline{0.338} & 0.471 & 0.290 & 0.495 & 0.586 & 0.342 & 0.532 & 0.668 & 0.347 & 0.503 & 0.617  \\
        PLIP    & 0.367 & 0.526 & 0.621 & 0.187 & 0.336 & 0.469 & 0.253 & 0.414 & 0.527 & 0.356 & 0.536 & 0.665 & 0.370 & 0.539 & \underline{0.650}  \\
        \midrule
        BLEEP & \underline{0.419} & \underline{0.554} & \underline{0.630} & 0.152 & 0.182 & 0.212 & \underline{0.318} & \underline{0.500} & \underline{0.614} & \textbf{0.443} & \textbf{0.603} & \textbf{0.704} & \underline{0.426} & \underline{0.550} & 0.623 \\
        OmiCLIP & 0.281 & 0.453 & 0.596 & 0.177 & 0.320 & \underline{0.485} & 0.231 & 0.382 & 0.532 & 0.342 & 0.520 & 0.636 & 0.329 & 0.502 & 0.605 \\
        \cmidrule(lr){1-16}
        \rowcolor{yellow!20} \textsc{Sigmma} 
        & \textbf{0.590} & \textbf{0.728} & \textbf{0.826} 
        & \textbf{0.402} & \textbf{0.630} & \textbf{0.813}
        & \textbf{0.333} & \textbf{0.559} & \textbf{0.731} 
        & \underline{0.394} & \underline{0.570} & \underline{0.687}
        & \textbf{0.472} & \textbf{0.591} & \textbf{0.687}\\
        \bottomrule
    \end{tabular}
    \vspace{0.2em}
    \begin{tabular}{c|ccc|ccc|ccc|ccc|ccc}
        \toprule
        & \multicolumn{15}{c}{ST $\rightarrow$ HE} \\
        \cmidrule(lr){2-16}
        Dataset 
        & \multicolumn{3}{c|}{\textit{HEST1k-LUAD}} 
        & \multicolumn{3}{c|}{\textit{HEST1k-PAAD}} 
        & \multicolumn{3}{c|}{\textit{HEST1k-SKCM}} 
        & \multicolumn{3}{c|}{\textit{HEST1k-IDC}} 
        & \multicolumn{3}{c}{\textit{in-house skin}} \\
        \cmidrule(lr){2-16}
        Model 
        & R@5\% & R@10\% & R@15\% 
        & R@5\% & R@10\% & R@15\% 
        & R@5\% & R@10\% & R@15\% 
        & R@5\% & R@10\% & R@15\% 
        & R@5\% & R@1\%0 & R@15\% \\
        \midrule[1.2pt]
        CLIP    & 0.297 & 0.413 & 0.526 & 0.141 & 0.284 & 0.390 & 0.285 & 0.473 & \underline{0.591} & \underline{0.445} & \textbf{0.675} & \textbf{0.798} & 0.354 & 0.513 & 0.619  \\
        PLIP    & 0.330 & 0.483 & 0.618 & \underline{0.213} & \underline{0.358} & \underline{0.475} & 0.274 & 0.435 & 0.543 & 0.440 & 0.639 & 0.767 & 0.371 & 0.552 & \underline{0.665} \\
        \midrule
        BLEEP & \underline{0.415} & \underline{0.568} & \underline{0.654} & 0.030 & 0.121 & 0.212 & \underline{0.330} & \underline{0.494} & 0.580 & \textbf{0.502} & \underline{0.655} & \underline{0.754} & \underline{0.419} & \underline{0.561} & 0.634 \\
        OmiCLIP & 0.281 & 0.501 & 0.599 & 0.165 & 0.318 & 0.435 & 0.242 & 0.403 & 0.495 & 0.412 & 0.612 & 0.742 & 0.335 & 0.514 & 0.632 \\
        \cmidrule(lr){1-16}
        \rowcolor{yellow!20}\textsc{Sigmma} 
        & \textbf{0.602} & \textbf{0.768} & \textbf{0.813}
        & \textbf{0.304} & \textbf{0.505} & \textbf{0.652}
        & \textbf{0.333} & \textbf{0.500} & \textbf{0.602}
        & 0.399 & 0.611 & 0.750 
        & \textbf{0.459} & \textbf{0.620} & \textbf{0.708}\\
        \bottomrule[1.2pt]
    \end{tabular}
    \vspace{-3pt}
    \label{table:crossmodal_retrieval}
\end{table*}
\begin{itemize}
    \item \textbf{Task 1. Gene expression prediction.} We perform linear probing on HE tile embeddings extracted from image encoders trained with \textsc{sigmma} and baseline models, following the HEST-1k benchmarking protocol \citep{jaume2024hest}. To prevent information leakage, the linear probe is trained and evaluated strictly on the training and test splits used during model training. Each HE tile embedding is reduced to 256 dimensions using PCA, followed by a simple ridge regression model trained to predict the expression levels of the top 50 highly variable genes. We report tile-level prediction performance as the mean $\pm$ standard deviation across tiles, using Pearson Correlation Coefficient (PCC) and Mean Squared Error (MSE) as evaluation metrics.
    \item \textbf{Task 2. Cross-modal retrieval} We report Recall@p\%, defined as the fraction of queries whose true counterpart, i.e., the HE–ST tile pair obtained from the same spatial location, appears within the top p\% (p=5, 10, 15) of retrieved candidates. The metric quantifies how accurately the model aligns HE and ST modalities at the tile level. The metric is evaluated on test tiles that were excluded during model training, ensuring a fair assessment.
\end{itemize}

\subsection{Task 1. Gene expression prediction} 
Here, our focus is to evaluate the quality of the learned image representation for gene expression prediction. Therefore, to avoid introducing biases from different methods' gene expression decoders, we use the image embedding output by each method and apply a ridge regression for gene expression prediction for each method.
Multi-modal alignment consistently enriches HE image embeddings by incorporating ST information (Suppl. Tab. \ref{table:improves_any_image_embeddings}). When applying \textsc{sigmma} on top of ResNet50 \citep{he2016deep}, H-Optimus-0 \citep{hoptimus0}, or UNI \citep{chen2024towards}, \textsc{Sigmma} improves representations across backbones, achieving up to 67\% lower MSE and 56\% higher PCC with UNI. Given its strong gains, we used UNI as the HE encoder backbone for all subsequent experiments.

We then compared \textsc{Sigmma} with existing baselines across five datasets (Tab. \ref{table:gex_prediction}). Across all datasets, \textsc{Sigmma} achieves the highest PCC. While \textsc{Sigmma} does not always obtain the lowest MSE, \textsc{Sigmma} consistently ranks among the top-performing models, highlighting its stable and robust performance across diverse tissue types.

\subsection{Task 2. Cross-modal retrieval}

Tab. \ref{table:crossmodal_retrieval} summarizes the cross-modal retrieval performance for both HE$\rightarrow$ST and ST$\rightarrow$HE. \textsc{Sigmma} delivers strong and consistent gains over existing baselines across most datasets, while overall performance on IDC remains relatively weak.
Overall, \textsc{Sigmma} achieves strong bi-directional alignment on most of the benchmark datasets.

\begin{figure}[b]
  \centering
  \includegraphics[width=1\linewidth]{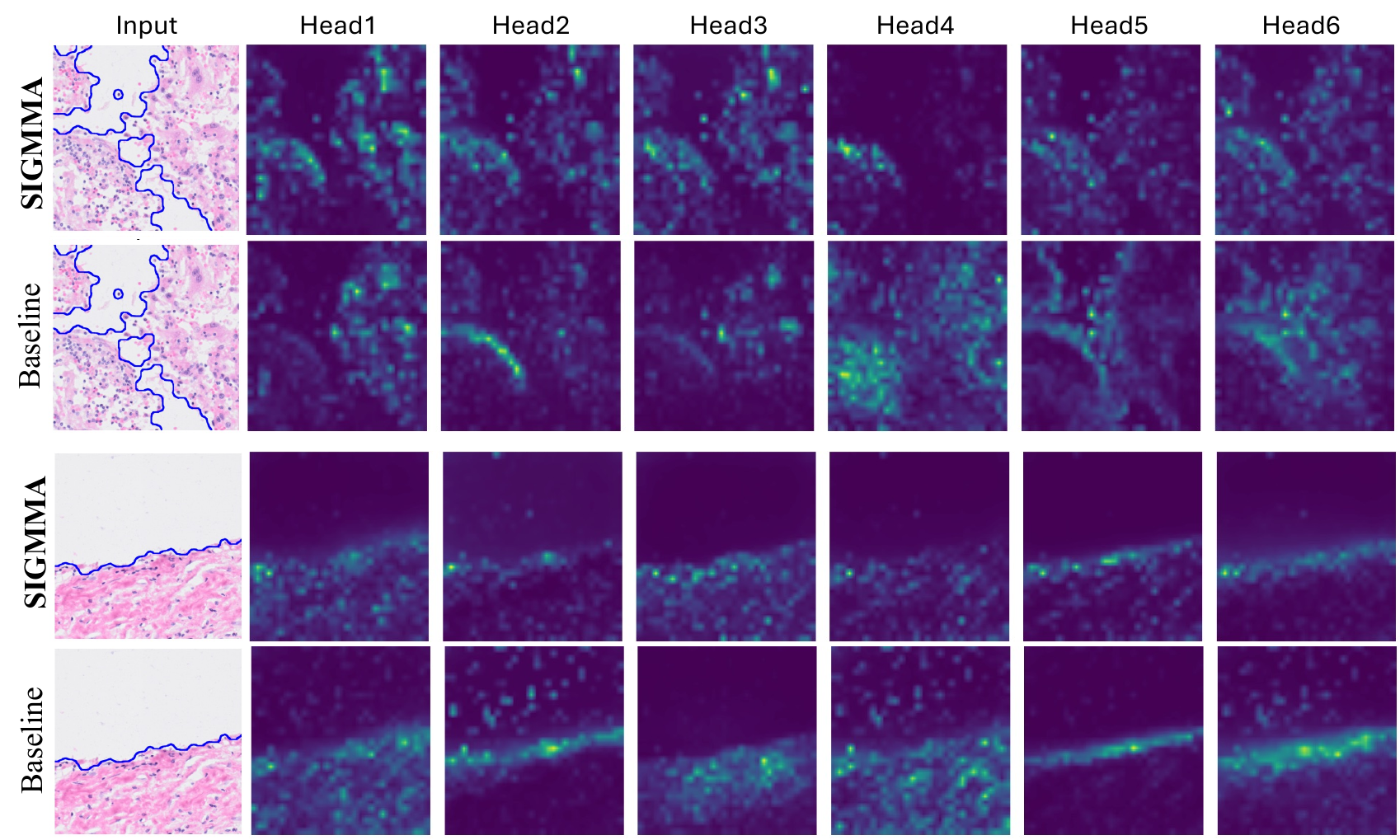}
  \vspace{-15pt} 
  \caption{Attention maps from six attention heads in the last encoder layer ($L = 24$) of the UNI image encoder backbone fine-tuned with \textsc{sigmma}, illustrating class-token–to-patch attention distributions. Blue contour overlaid on the input images indicates the cell-segmentation mask, marking the boundaries of cell-rich regions.}
  \label{fig:attention-map}
\end{figure}

\subsection{Qualitative evaluation}
\paragraph*{Cell-aware attention via \textsc{Sigmma}}
We analyzed the attention map of image encoders from \textsc{sigmma}. 
In HE, the white areas within tissue correspond to adipose regions where lipids get washed out during processing, so they appear as cell-sparse empty spaces (Fig. \ref{fig:attention-map}, row 1, Input). With \textsc{Sigmma}, the attention clearly focuses on nuclei-rich, cell-dense regions, showing sharp and localized activations around individual cells, which means it actually captures fine-grained cell morphology (Fig. \ref{fig:attention-map} row 1, Head 5-6). On the other hand, the  baseline image foundation model, UNI, tends to put more attention on tissue boundaries and adipose regions, which are cell-sparse areas (Fig. \ref{fig:attention-map} row 2, Head 5-6). Even though these regions don’t contain cells, they still stand out morphologically, so UNI attends to these coarse structural cues rather than true cell-level features. Similarly, \textsc{Sigmma} shows low attention scores in the out-of-tissue regions (Fig. \ref{fig:attention-map}, row 3–4, Head 2, 4, 6). Overall, these results demonstrate that \textsc{Sigmma} effectively shifts the model’s focus from coarse tissue structures to biologically meaningful, cell-level morphology.





\subsection{Ablation study}
As shown in Tab. \ref{table:ablation_task1}, we analyze the impact of the core components of \textsc{sigmma}: (1) cell graph, (2) multi-scale loss, and (3) graph sparsification via stochastic edge addition.
For the ablation of the cell graph, we replaced the spatial graph representation with a 1D sequence of genes.
For the ablation of multi-scale loss, we removed the micro- and meso-scale objectives and trained only with the macro-scale (single-scale) alignment loss.
For the ablation of graph sparsification module, instead of selectively sampling edges through stochastic addition, we connected all nodes within neighboring patches, resulting in a fully connected local graph.
Tab. \ref{table:ablation_task1} shows that each component contributes to performance gains: adding the multi-scale loss and graph sparsification progressively improves prediction accuracy, with all components combined achieving the lowest MSE and highest PCC.
The ablation study for Task 2 is provided in the Suppl. Sec. \ref{suppl:result}.
Overall, the ablation study shows that each component contributes to performance, with multi-scale loss and graph sparsification having the largest impact.
\begin{table}[t]
    \centering
    \scriptsize
    \caption{Ablation of core components of \textsc{sigmma} on HEST1k-LUAD dataset for gene expression prediction task.}
    \setlength{\tabcolsep}{6pt}
    \renewcommand{\arraystretch}{1.2}
    \begin{tabular}{ccc|cc}
        \toprule[1.2pt]
         \multicolumn{3}{c|}{Components} & \multicolumn{2}{c}{Task 1.} \\
        \cmidrule(lr){1-3}\cmidrule(lr){4-5}
         \makecell{Cell graph} & \makecell{Multi-scale \\ loss} & \makecell{Graph \\ sparsification} & MSE (↓) & PCC (↑) \\
        \midrule[1.2pt]
           &  &  & 0.032$\pm$0.018 & 0.345$\pm$0.035  \\
         \checkmark &  &  & 0.039$\pm$0.018  &  0.268$\pm$0.032 \\
         \checkmark & \checkmark &  & 0.020$\pm$0.014  &  0.645$\pm$0.046 \\
         \rowcolor{yellow!20} \checkmark & \checkmark & \checkmark & \textbf{0.015$\pm$0.007} & \textbf{0.741$\pm$0.023} \\
        \bottomrule[1.2pt]
    \end{tabular}
    \label{table:ablation_task1}
    \vspace{-3pt}
\end{table}

\subsection{Biological application}
\label{subsection:biological_application}
\begin{figure*}[t]
  \centering
  \includegraphics[width=0.80\linewidth]{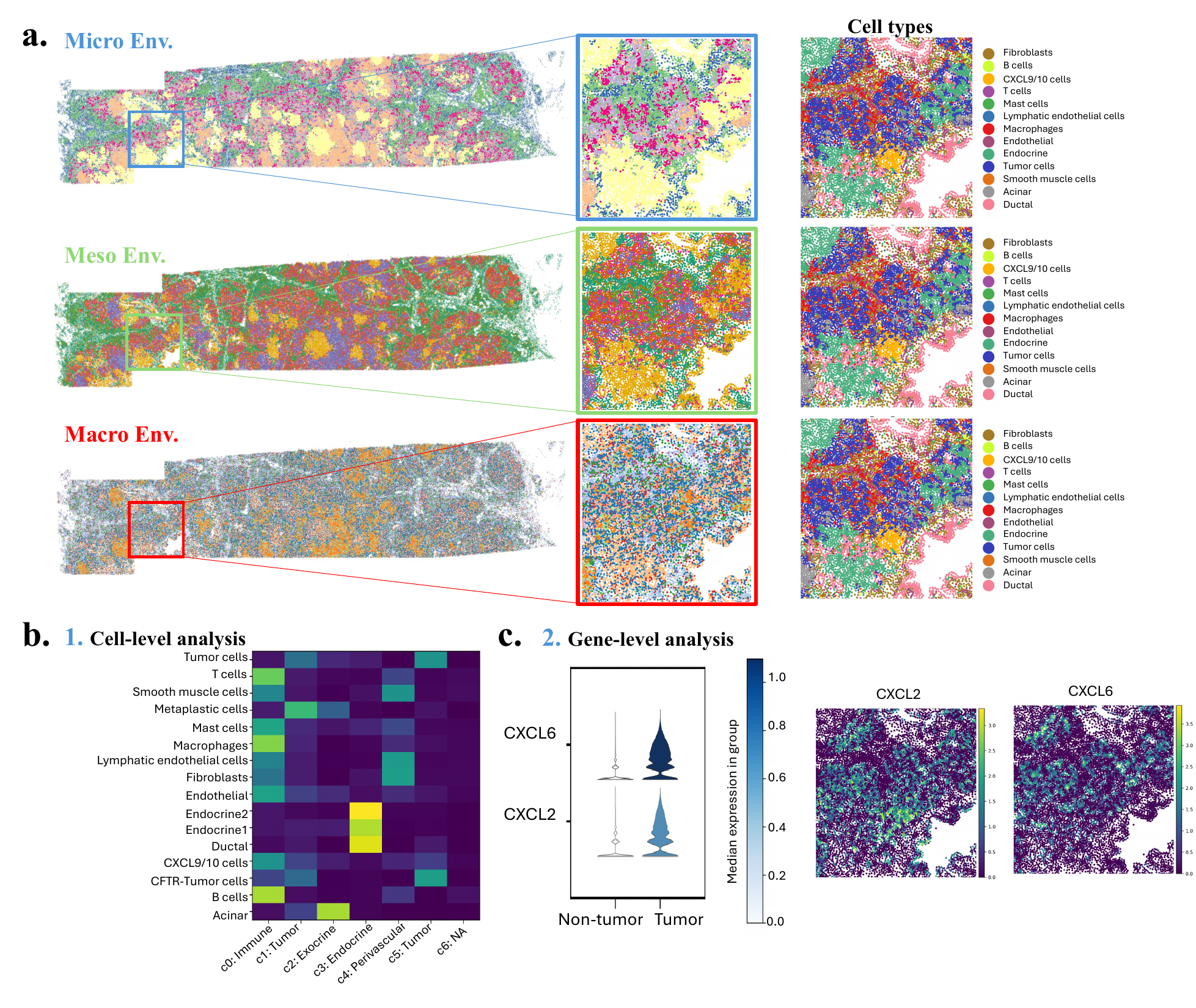}
  \vspace{-15pt}
  \caption{\textbf{Biological interpretation of \textsc{Sigmma} cell-level based embeddings}. \textbf{(a)} Left: Microenvironment clusters projected onto the 2D spatial map across scales for the whole slide. Middle: Close-up view of the tissue boundary highlighting separation between tumor and non-tumor regions. Right: Ground-truth cell-type annotation for comparison. \textbf{(b)} Heatmap showing cell-type proportions for each microenvironment at the micro scale (x-axis: cell clusters, y-axis: cell types). \textbf{(c)} Left: Violin plots of tumor-promoting genes (\textit{CXCL2}, \textit{CXCL6}) illustrating their expression distributions in aggregated tumor versus non-tumor microenvironments. Right: Spatial projection of these gene expression patterns onto the 2D map for the region of interest shown in (a, middle).}
  \vspace{-2pt}
  \label{fig:bio-interpretation}
\end{figure*}

\paragraph{Multi-scale cell embeddings from \textsc{sigmma} to the pancreatic tumor microenvironment} We next evaluated whether the multiscale cell-level embeddings learned by \textsc{sigmma} capture meaningful biological structure in a publicly available section from PAAD tissue. For more details on the section, see Suppl. Sec. \ref{suppl:data}, which includes a full description of the dataset composition and sources.
We performed clustering of \textsc{sigmma} embeddings at the micro, meso, and macro-scale and projected these clusters back onto a PAAD tissue section, comparing them with reference cell labels (Fig. \ref{fig:bio-interpretation}a). Across all three scales, the inferred microenvironments formed coherent and spatially contiguous domains. Most notably at the micro- and meso-scales, \textsc{sigmma} cleanly delineated tumor nests from surrounding non-tumor tissue, despite the absence of any supervision from cell-type labels.
To characterize the cellular context represented by these embeddings, we quantified the cell-type composition of each microenvironment cluster, focusing on the micro-scale (Fig. \ref{fig:bio-interpretation}b). We identified six resolvable microenvironments. Two microenvironments were composed predominantly of tumor cells. A third represented an inflammatory infiltrate enriched for multiple immune cell types, including T cells and B cells, that were spatially excluded from the tumor region. Three additional microenvironments corresponding to known pancreatic structures, including perivascular, endocrine, and exocrine compartments. These patterns are consistent with the expected organization of pancreatic tumor tissue and indicate that \textsc{sigmma} learns fine-grained microenvironmental structure directly from spatial molecular context.
Finally, we asked whether the learned embeddings capture relevant molecular signatures. Differential expression analysis between tumor-associated microenvironments and the immune microenvironment excluded from the tumor identified two chemokines, \textit{CXCL2} and \textit{CXCL6}, both implicated in the recruitment of anti-inflammatory and immunosuppressive myeloid cells that can mediate T cell and B cell exclusion in other cancer contexts \citep{hu2021regulation}. Spatial projection of \textit{CXCL2} and \textit{CXCL6} expression confirmed that both signals localize sharply to tumor regions in the PAAD section, consistent with a tumor-specific chemokine program (Fig. \ref{fig:bio-interpretation}c),
Overall, these results show that \textsc{sigmma} not only recovers structural hallmarks of immune exclusion in PAAD but also resolves molecular features that may contribute to the establishment of immunosuppressive cell states within the tumor microenvironment. Thus, \textsc{sigmma} captures biologically interpretable tissue organization across scales and reveals spatially coherent molecular programs that align with the underlying architecture of pancreatic cancer.

\section{Conclusion}
In this work, we presented \textsc{Sigmma}, a hierarchical multi-modal alignment framework that learns joint representations of HE and ST across micro, meso, and macro scales. Although multi-scale contrastive learning has been explored in other domains, \textsc{Sigmma} is the first to address the graph receptive field–ROI mismatch that uniquely arises in cell-resolution ST. \textsc{Sigmma} effectively captures both fine-grained cellular structure and broader tissue context. 





\section{Author contribution}
D.Jeong led the project, proposed the methodology, and designed and executed the core experiments.
AV co-designed the experiments, provided project feedback, and supported project management.
CRS handled data preprocessing and D.Jafree contributed to the biological interpretation.
MM, KL, and MV assisted with baseline benchmark experiments.
SB contributed to the design and execution of an evaluation task.
YC and AB developed and ran the data preprocessing pipeline.
LS, VBMS, and ARF generated the datasets.
MH and ML supervised the project.

{
    \small
    \bibliographystyle{ieeenat_fullname}
    \bibliography{main}
}

\clearpage
\setcounter{page}{1}
\appendix
\maketitlesupplementary

Details on the dataset are provided in Sec. \ref{suppl:data}, and methodological descriptions are included in Sec. \ref{suppl:method}. Additional results that could not be presented in the main text due to space constraints are shown in Sec. \ref{suppl:result}. Finally, Sec. \ref{suppl:discussion} presents an analysis of challenging cases, limitations, and directions for future work.

\section{Data}
\label{suppl:data}

\subsection{Data acquisition}
\label{suppl:data_acquisition}
\paragraph{HEST1k dataset.} 
The HEST1k dataset~\citep{jaume2024transcriptomics} is a publicly available benchmark comprising paired HE image and Xenium-ST data, accessible at \url{https://huggingface.co/datasets/MahmoodLab/hest}.
Table~\ref{table:hest1k_dataset_summary} provides a comprehensive summary of all Xenium sections within HEST1k, including tissue types, cell counts, and the presence of cell-type annotations. All sections in the table have paired post-Xenium HE morphology images.
Cell-type annotations were derived from the \textit{TENX116} section, which serves as the reference dataset for the biological analyses presented in Sec .~\ref {subsection:biological_application}.

\begin{table}[!h]
    \centering
    \scriptsize
    \caption{Summary of HEST1k HE-Xenium ST sections.}
    \vspace{-2mm}
    \begin{tabular}{c|c|c|c|c}
    \toprule[1.2pt]
     Tissue type & Section ID & \# of cells & \makecell{Cell-type \\ annotations} & Source \\
    \midrule[1.2pt]
     \multirow{2}{*}{LUAD} 
        & TENX118 & 162,254 &  & \href{https://www.10xgenomics.com/datasets/preview-data-ffpe-human-lung-cancer-with-xenium-multimodal-cell-segmentation-1-standard}{Public} \\
        & TENX141 & 161,000 &  & \href{https://www.10xgenomics.com/datasets/ffpe-human-lung-cancer-data-with-human-immuno-oncology-profiling-panel-and-custom-add-on-1-standard}{Public} \\
    \midrule
     \multirow{3}{*}{PAAD} 
        & TENX116 & 190,965 & O & \href{https://www.10xgenomics.com/datasets/pancreatic-cancer-with-xenium-human-multi-tissue-and-cancer-panel-1-standard}{Public} \\
        & TENX140 & 235,099 &  & \href{https://www.10xgenomics.com/datasets/ffpe-human-ductal-adenocarcinoma-data-with-human-immuno-oncology-profiling-panel-1-standard}{Public} \\
        & TENX126 & 140,702 &  & \href{https://www.10xgenomics.com/datasets/ffpe-human-pancreas-with-xenium-multimodal-cell-segmentation-1-standard}{Public} \\
    \midrule
     \multirow{2}{*}{SKCM}
        & TENX115 & 106,980 &  & \href{https://www.10xgenomics.com/datasets/human-skin-preview-data-xenium-human-skin-gene-expression-panel-1-standard}{Public} \\
        & TENX117 & 87,499 &  & \href{https://www.10xgenomics.com/datasets/human-skin-preview-data-xenium-human-skin-gene-expression-panel-add-on-1-standard}{Public} \\
    \midrule
     \multirow{4}{*}{IDC}
        & TENX95  & 574,852 &  & \href{https://www.10xgenomics.com/datasets/ffpe-human-breast-with-pre-designed-panel-1-standard}{Public} \\
        & TENX99  & 892,966 &  & \href{https://www.10xgenomics.com/datasets/ffpe-human-breast-using-the-entire-sample-area-1-standard}{Public} \\
        & NCBI783 & 142,272 &  & \href{https://www.10xgenomics.com/products/xenium-in-situ/preview-dataset-human-breast}{Public} \\
        & NCBI785 & 167,780 &  & \href{https://www.10xgenomics.com/products/xenium-in-situ/preview-dataset-human-breast}{Public} \\
    \bottomrule[1.2pt]
    \end{tabular}
    \label{table:hest1k_dataset_summary}
\end{table}

\paragraph{In-house dataset.}
The in-house dataset consists of 21 Xenium ST sections derived from 13 patients with either eczema or skin warts. Eczema samples were profiled using the Xenium Prime 5K Human Pan-Tissue \& Pathways Panel, whereas skin wart samples were processed with the Xenium Immuno-Oncology Panel. Tissue sections were placed into the active capture area on Xenium slides and stored accordingly until being processed for in situ gene expression according to the manufacturer’s protocol. All Xenium slides were processed using the Xenium Prime 5K Human Pan Tissue \& Pathways Panel, including the standard cell segmentation antibody staining.  Due to confidentiality, detailed section-level metadata cannot be shared; however, we report an average cell count of approximately 16,704$\pm$6,693 cells per section.

\subsection{Data preprocessing}
\label{suppl:data_preprocessing}
\paragraph{HE image.} Since the HE image and Xenium ST profile are acquired on different imaging systems, their spatial coordinates are not directly comparable. To place morphological information from the HE image into the same spatial coordinate system, whole-slide HE image and immunofluorescence (IF) were registered using Palom \citep{Chen_PALOM_-_Piecewise_2025}, a piecewise registration framework for layers of mosaics. Image registration was performed using the DAPI (IF) and green (HE) channels. Coarse affine alignment was initialized using 4,000 keypoints, followed by local shift refinement and constraint optimization. The final HE image was reconstructed using a blockwise affine transformation and rescaled to $0.5 \mu \text{m}$/pixel at 20x magnification level. The Public HEST1k dataset was provided with registration already completed. All sections are tessellated into $256 \times 256$ pixel-sized tiles.

\paragraph{ST Xenium profile.}
All Xenium ST data is processed with the 10X platform \citep{janesick2023high}, which provides built-in cell segmentation and outputs cell-resolved features by default. We extracted ST tiles corresponding to the same spatial region as each HE tile and constructed tile-level representations from the transcriptome profiles of all cells within each tile. For graph representation of each ST tile, we constructed cell graphs using \texttt{squidpy.gr.spatial\_neighbors} with default parameters \citep{squidpy_spatial_neighbors}, which builds a 6-nearest-neighbor graph from Euclidean coordinates to capture local spatial proximity among cells. For sequence representation of ST in the VL model, we followed Loki preprocessing pipeline \citep{chen2025visual}, selecting the top 50 genes by tile-level mean expression and ordering the gene names in descending mean expression.
\subsection{Data splitting strategy}
\begin{figure}[h]
  \centering
    \includegraphics[width=1\linewidth]{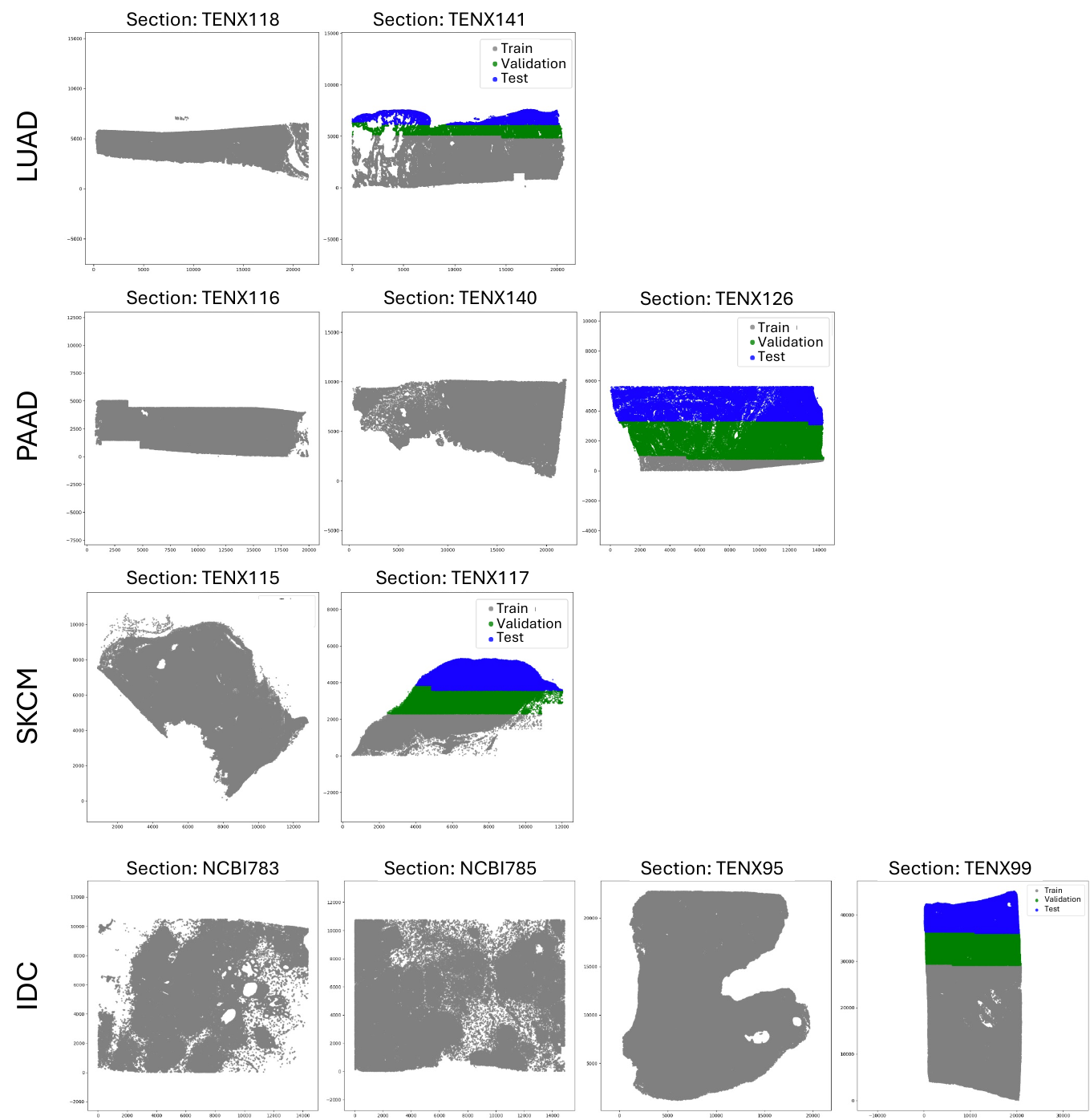}
    \caption{Train-validation-test split of HEST1k dataset.}\
    \vspace{-10pt}
    \label{fig:viz_split}
\end{figure}

\paragraph{HEST1k Dataset.} 
Due to the limited number of paired HE–Xenium ST sections in public datasets (Tab. \ref{table:hest1k_dataset_summary}), section-level split is not feasible, as it would not provide enough data for stable model training. Additionally, assigning spatially adjacent tiles to different splits can lead to spatial leakage across data subsets, since such tiles often share morphological and molecular characteristics. To address these issues, we employ a \textit{spatially-stratified tile split} for the HEST1k dataset, assigning validation and test tiles to non-overlapping spatial regions (Fig. \ref{fig:viz_split}). All splits use an 8:1:1 ratio for training, validation, and testing.

\paragraph{In-house dataset.} Given the larger number of available sections, we employ a section-level split for the in-house dataset. Each section is assigned entirely to one subset to avoid information leakage across data splits. The dataset is randomly partitioned into training, validation, and test sets with an 8:1:1 ratio.

\section{Method}
\label{suppl:method}

\subsection{Implementation detail}
We use UNI \citep{chen2024towards} as the HE encoder backbone, which is a ViT-L/16 architecture. For each HE tile, we generate multi-scale crops and resize each patch to the ViT input resolution, 224 $\times$ 224 pixels. Each patch is encoded using the UNI model, and the CLS token output is used as its patch-level embedding. Patch embeddings within each scale are then mean-pooled and L2-normalized to produce a single representation per scale. This procedure is applied from the micro to macro scale, yielding a set of multi-scale HE embeddings.

For the ST encoder. node embeddings are initialized with STEMO features \citep{seb2025STEMO}, followed by SAGEConv\citep{hamilton2017inductive}-based message passing and stochastic edge addition layer \citep{piao2022sparse}. All GNN components are implemented using the \texttt{dgl} library. Graph-level embeddings are then obtained via attention pooling. This procedure is applied hierarchically from micro to macro scales to produce multi-scale ST representations.

At each scale, HE and ST embeddings are passed through simple MLP projection heads to obtain vectors of the same dimensionality, and a symmetric InfoNCE loss is then applied to align the modalities. The hidden dimensions for HE and ST embeddings are set to 256 or 512. All projection and encoder layers use LeakyReLU activations.

We optimize the model using AdamW with a StepLR scheduler and weight decay. To obtain a larger effective batch size, which is essential for contrastive learning, we employ 10 steps of gradient accumulation with a batch size of 128 and incorporate cross-batch negatives. All experiments are conducted on a single A100 SXM4 GPU with 80GB RAM. 

The hyperparameters were tuned by grid search, as a combination of learning rate [0.001, 0.0001, 0.00001, 0.000001], weight decay value [0.001, 0.0001, 0.00001], dropout rate [0.1, 0.2, 0.3, 0.4, 0.5], and feature dimension of HE/ST embeddings [256, 512]. Each hyperparameter combination was fed into the model, and the validation loss was calculated. The model checkpoint with the smallest evaluation loss is saved for testing. You can find the optimized hyperparameter settings in the code under \textit{config/\{dataset\}\_\{model\_name\}.yaml}.



\subsection{Evaluation metrics.}

\paragraph{Gene expression prediction.}
For each tile, we assess how accurately the model predicts the expression vector over the top $G$ (G=50) highly variable genes. Let $y_{i,g}$ and $\hat{y}_{i,g}$ denote the ground-truth and predicted expression values of gene $g$ in tile $i$. Tile-level MSE and PCC are computed across genes as follows:
\begin{align}
    \text{MSE}_i &= \frac{1}{G} \sum_{g=1}^{G} (y_{i,g} - \hat{y}_{i,g})^2 \\[4pt]
    \text{PCC}_i &=
        \frac{
            (\mathbf{y}_i - \bar{\mathbf{y}}_i)^\top
            (\hat{\mathbf{y}}_i - \bar{\hat{\mathbf{y}}}_i)
        }{
            \| \mathbf{y}_i - \bar{\mathbf{y}}_i \|_2 \;
            \| \hat{\mathbf{y}}_i - \bar{\hat{\mathbf{y}}}_i \|_2
        }
\end{align}

We report dataset-level performance as the mean~$\pm$~standard deviation of 
$\text{MSE}_i$ and $\text{PCC}_i$ across all tiles.

\paragraph{Cross-modal retrieval.}
For a query tile $q$ from one modality (HE or ST), we rank all tiles in the other modality by embedding similarity. Retrieval accuracy is measured by whether the paired tile appears within the top-$p$\%. Recall@$p$\% is defined as follows:
\begin{equation}
    \text{Recall@}p\% 
    = \frac{1}{N} \sum_{q=1}^{N} 
    \mathbbm{1}\!\left( y_q \in \text{TopK}(q) \right),
\end{equation}
where $N$ denotes the number of query tiles, $K = \lfloor p\% \times N \rfloor$ and $\mathbbm{1}(\cdot)$ is the indicator function.

\subsection{Fine-tuning of baselines.}
All baselines were either fine-tuned on our standardized data splits or evaluated using their official checkpoints. Each model was fine-tuned using the loss function specified in the original implementation. For BLEEP, no pretrained BLEEP checkpoint is provided, and only the ResNet50 \citep{he2016deep} backbone is publicly available; therefore, we initialized BLEEP with this backbone and fine-tuned the model on our dataset to ensure a consistent and fair comparison.

\section{Result}
\label{suppl:result}

\subsection{Ablation study}
Here, we present ablation results for the core components of the cross-retrieval task (Task 2) on the HEST1k-LUAD dataset. As shown in Tab.~\ref{table:ablation_task2}, each component contributes to performance gains in the retrieval task: adding the multi-scale loss and graph sparsification progressively improves retrieval accuracy, and integrating all components yields the best overall performance.
\begin{table}[!h]
    \centering
    \scriptsize
    \caption{Ablation of core components of \textsc{sigmma} on HEST1k-LUAD dataset for the cross-modal retrieval task.}
    \setlength{\tabcolsep}{6pt}
    \renewcommand{\arraystretch}{1.2}
    \begin{tabular}{ccc|ccc}
        \toprule[1.2pt]
         \multicolumn{3}{c|}{Components} & \multicolumn{3}{c}{Task 2. HE $\rightarrow$ ST} \\
        \cmidrule(lr){1-3}\cmidrule(lr){4-6}
         \makecell{Cell graph} & \makecell{Multi-scale \\ loss} & \makecell{Graph \\ sparsification} & R@5\% & R@10\% & R@15\%\\
        \midrule[1.2pt]
           &  &  & 0.517 & 0.694 & 0.768 \\
         \checkmark &  &  & 0.480 & 0.639 & 0.737 \\
         \checkmark & \checkmark &  & 0.550 & 0.667 & 0.786 \\
         \rowcolor{yellow!20} \checkmark & \checkmark & \checkmark & \textbf{0.590} & \textbf{0.728} & \textbf{0.826} \\
        \bottomrule[1.2pt]
    \end{tabular}
    \begin{tabular}{ccc|ccc}
        \toprule[1.2pt]
         \multicolumn{3}{c|}{Components} & \multicolumn{3}{c}{Task 2. ST $\rightarrow$ HE} \\
        \cmidrule(lr){1-3}\cmidrule(lr){4-6}
         \makecell{Cell graph} & \makecell{Multi-scale \\ loss} & \makecell{Graph \\ sparsification} & R@5\% & R@10\% & R@15\%\\
        \midrule[1.2pt]
           &  &  & 0.529 & 0.673 & 0.780 \\
         \checkmark &  &  & 0.459 & 0.621 & 0.722 \\
         \checkmark & \checkmark &  & 0.514 & 0.685 & 0.761 \\
         \rowcolor{yellow!20} \checkmark & \checkmark & \checkmark & \textbf{0.602} & \textbf{0.768} & \textbf{0.813} \\
        \bottomrule[1.2pt]
    \end{tabular}
    \label{table:ablation_task2}
    \vspace{-3pt}
\end{table}

Extending the ablation in the main manuscript, we additionally explored (1) different HE image backbones, ResNet50 \citep{he2016deep}, H-Optimus-0 \citep{hoptimus0}, and UNI \citep{chen2024towards}, and (2) different ST backbones for cell embedding initialization, Harmony \citep{korsunsky2019fast}, Novae \citep{blampeyNovae2024}, and STEMO \citep{seb2025STEMO}. These experiments assess the impact of backbone choices and support the architectural decisions made in our final model. \textsc{Sigmma}, built upon the UNI vision backbone, consistently achieves the best performance across both downstream tasks (Tab. \ref{table:improves_any_image_embeddings}, Tab. \ref{table:improves_any_image_embeddings_retrieval}).
On the other hand, among the ST backbones, STEMO tended to perform well overall across both downstream tasks (Tab. \ref{table:ablation_STbackbone_task1}, Tab. \ref{table:ablation_STbackbone_task2}). Accordingly, these results justify selecting UNI and STEMO as the vision and ST backbones in our framework.
\begin{table}[!h]
    \centering
    \scriptsize
    \setlength{\tabcolsep}{6pt}
    \caption{Ablation of vision backbone in HEST1k-LUAD dataset for task 1.}
    \vspace{-2mm}
    \begin{tabular}{c|cc}
    \toprule[1.2pt]
     & \multicolumn{2}{c}{Task 1. Gene expression prediction.} \\
    \cmidrule(lr){2-3}
     Model & MSE (↓) & PCC (↑) \\
    \midrule[1.2pt]
      ResNet50             & 0.052$\pm$0.047 & 0.365$\pm$0.079\\
      \text{\textsc{Sigmma} (ResNet50)}            & 0.031$\pm$ 0.035 & 0.389$\pm$0.064 \\
    \midrule
      H-Optimus-0           & 0.035$\pm$0.034 & 0.512$\pm$0.078 \\
      \text{\textsc{Sigmma} (H-Optimus-0)}         & 0.020$\pm$0.018 & 0.673$\pm$0.030 \\
    \midrule
      UNI             & 0.046$\pm$0.041 & 0.476$\pm$0.064 \\
      \text{\textsc{Sigmma} (UNI)}        & \textbf{0.015$\pm$0.007} &  \textbf{0.741 $\pm$0.023} \\
    \bottomrule[1.2pt]
    \end{tabular}
    \label{table:improves_any_image_embeddings}
\end{table}
\begin{table}[!h]
    \centering
    \scriptsize
    \setlength{\tabcolsep}{1pt}
    \caption{Ablation of vision backbones in HEST1k-LUAD dataset for task 2.}
    \begin{tabular}{c|ccc|ccc}
    \toprule[1.2pt]
     & \multicolumn{6}{c}{Task 2. Cross-modal retrieval.} \\
     \cmidrule(lr){2-7}
     & \multicolumn{3}{c|}{HE $\rightarrow$ ST} & \multicolumn{3}{c}{ST $\rightarrow$ HE}\\
    \cmidrule(lr){2-7}
     Model & R@5\% & R@10\% & R@15\% & R@5\% & R@10\% & R@15\%\\
    \midrule[1.2pt]
      \text{\textsc{Sigmma} (ResNet50)}       & 0.086 & 0.199 & 0.260 & 0.135 & 0.229 & 0.333 \\
    \midrule
      \text{\textsc{Sigmma} (H-Optimus-0)}    & 0.563 & 0.676 & 0.777 & 0.554 & 0.688 & 0.774\\
    \midrule
      \text{\textsc{Sigmma} (UNI)}            & \textbf{0.590} & \textbf{0.728} & \textbf{0.826} & \textbf{0.602} & \textbf{0.768} & \textbf{0.813} \\
    \bottomrule[1.2pt]
    \end{tabular}
    \label{table:improves_any_image_embeddings_retrieval}
\end{table}

\begin{table}[!h]
    \centering
    \scriptsize
    \setlength{\tabcolsep}{6pt}
    \caption{Ablation of ST backbone in HEST1k-LUAD dataset for task 1.}
    \vspace{-2mm}
    \begin{tabular}{c|cc}
    \toprule[1.2pt]
     & \multicolumn{2}{c}{Task 1. Gene expression prediction.} \\
    \cmidrule(lr){2-3}
     Model & MSE (↓) & PCC (↑) \\
    \midrule[1.2pt]
      \text{\textsc{Sigmma} (Harmony)}       & \textbf{0.005$\pm$0.005} & 0.498$\pm$0.076 \\
    \midrule
      \text{\textsc{Sigmma} (Novae)}         & 0.011$\pm$0.003 & 0.606$\pm$0.048 \\
    \midrule
      \text{\textsc{Sigmma} (STEMO)}         & 0.015$\pm$0.007 & \textbf{0.741$\pm$0.023}  \\
    \bottomrule[1.2pt]
    \end{tabular}
    \label{table:ablation_STbackbone_task1}
\end{table}
\begin{table}[!h]
    \centering
    \scriptsize
    \setlength{\tabcolsep}{1pt}
    \caption{Ablation of ST backbone in HEST1k-LUAD dataset for task 2.}
    \begin{tabular}{c|ccc|ccc}
    \toprule[1.2pt]
     & \multicolumn{6}{c}{Task 2. Cross-modal retrieval.} \\
     \cmidrule(lr){2-7}
     & \multicolumn{3}{c|}{HE $\rightarrow$ ST} & \multicolumn{3}{c}{ST $\rightarrow$ HE}\\
    \cmidrule(lr){2-7}
     Model & R@5\% & R@10\% & R@15\% & R@5\% & R@10\% & R@15\%\\
    \midrule[1.2pt]
      \text{\textsc{Sigmma} (Harmony)}  & 0.529 & 0.723 & 0.820 & 0.526 & 0.730 & \textbf{0.830} \\
    \midrule
      \text{\textsc{Sigmma} (Novae)}    & 0.039 & 0.086 & 0.148 & 0.049 & 0.099 & 0.148 \\
    \midrule
      \text{\textsc{Sigmma} (STEMO)}            & \textbf{0.590} & \textbf{0.728} & \textbf{0.826} & \textbf{0.602} & \textbf{0.768} & 0.813 \\
    \bottomrule[1.2pt]
    \end{tabular}
    \label{table:ablation_STbackbone_task2}
\end{table}

\subsection{Sensitivity analysis on tile size}
We further conduct a sensitivity analysis on tile size, as tile resolution is a common source of variability in histopathology pipelines. Tile size 224 is a common choice in general vision models, while 256 is the standard image size used in many histopathology foundation models \citep{chen2024towards, vorontsov2024foundation, hoptimus0}. As shown in Tab. \ref{table:sensitivity_task1} and Tab. \ref{table:sensitivity_task2}, performance varies moderately across 224, 256, and 512 px tiles, but \textsc{sigmma} shows no collapse or strong dependence on any specific configuration. Mid-sized tiles (224/256 px) perform well across both tasks, while 512 px tiles show reduced performance, potentially due to increased heterogeneity within larger regions. Overall, this sensitivity analysis shows that while performance varies with tile resolution, \textsc{sigmma} remains generally robust, and medium tile sizes tend to provide favorable performance across tasks.

\begin{table}[!h]
    \centering
    \scriptsize
    \setlength{\tabcolsep}{6pt}
    \caption{Sensitivity analysis for task 1.}
    \vspace{-2mm}
    \begin{tabular}{c|cc}
    \toprule[1.2pt]
     & \multicolumn{2}{c}{Task 1. Gene expression prediction.} \\
    \cmidrule(lr){2-3}
     Model & MSE (↓) & PCC (↑) \\
    \midrule[1.2pt]
      \text{224 $\times$ 224} & 0.012$\pm$0.005 & 0.677$\pm$0.020 \\
    \midrule
      \text{256 $\times$ 256} & 0.015$\pm$0.007 & 0.741$\pm$0.023 \\
    \midrule
      \text{512 $\times$ 512} & 0.011$\pm$0.004 & 0.438$\pm$0.065 \\
    \bottomrule[1.2pt]
    \end{tabular}
    \label{table:sensitivity_task1}
\end{table}
\begin{table}[!h]
    \centering
    \scriptsize
    \setlength{\tabcolsep}{1pt}
    \caption{Sensitivity analysis for task 2.}
    \begin{tabular}{c|ccc|ccc}
    \toprule[1.2pt]
     & \multicolumn{6}{c}{Task 2. Cross-modal retrieval.} \\
     \cmidrule(lr){2-7}
     & \multicolumn{3}{c|}{HE $\rightarrow$ ST} & \multicolumn{3}{c}{ST $\rightarrow$ HE}\\
    \cmidrule(lr){2-7}
     Model & R@5\% & R@10\% & R@15\% & R@5\% & R@10\% & R@15\%\\
    \midrule[1.2pt]
      \text{224 $\times$ 224}  & 0.616 & 0.755 & 0.817 & 0.598 & 0.744 & 0.794 \\
    \midrule
      \text{256 $\times$ 256}  & 0.590 & 0.728 & 0.826 & 0.602 & 0.768 & 0.813  \\
    \midrule
      \text{512 $\times$ 512}  & 0.453 & 0.605 & 0.721 & 0.523 & 0.640 & 0.733 \\
    \bottomrule[1.2pt]
    \end{tabular}
    \label{table:sensitivity_task2}
\end{table}

\section{Discussions}
\label{suppl:discussion}

\subsection{Analysis of challenging cases}
\label{subsec:inspection_on_failure_case}

\begin{figure}[h]
  \centering
  \includegraphics[width=0.99\linewidth]{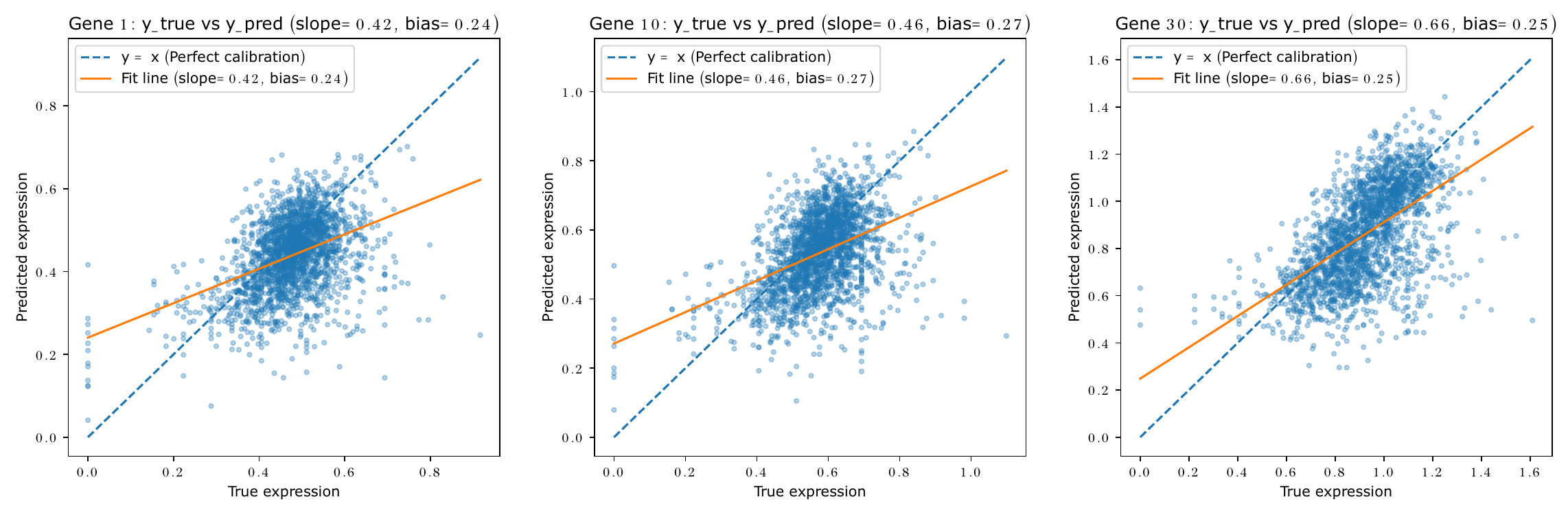}
  \caption{Calibration analysis of gene expression predictions.} 
  \vspace{-10pt}
  \label{fig:calibration}
\end{figure}

\paragraph{Understanding the PCC–MSE discrepancy.}
Although \textsc{sigmma} achieves high PCC across genes, indicating that it accurately captures the relative variation of expression across tiles, inspection of calibration plots reveals a consistent miscalibration in absolute prediction values. As shown in Fig. \ref{fig:calibration}, the regression line has slopes $<$ 1 and positive intercepts, indicating that the model underestimates variation while introducing a systematic bias. This global calibration mismatch increases MSE despite preserving rank-order consistency, explaining the discrepancy between PCC and MSE observed in Tab. \ref{table:gex_prediction}.

\begin{figure}[h]
  \centering
  \includegraphics[width=0.99\linewidth]{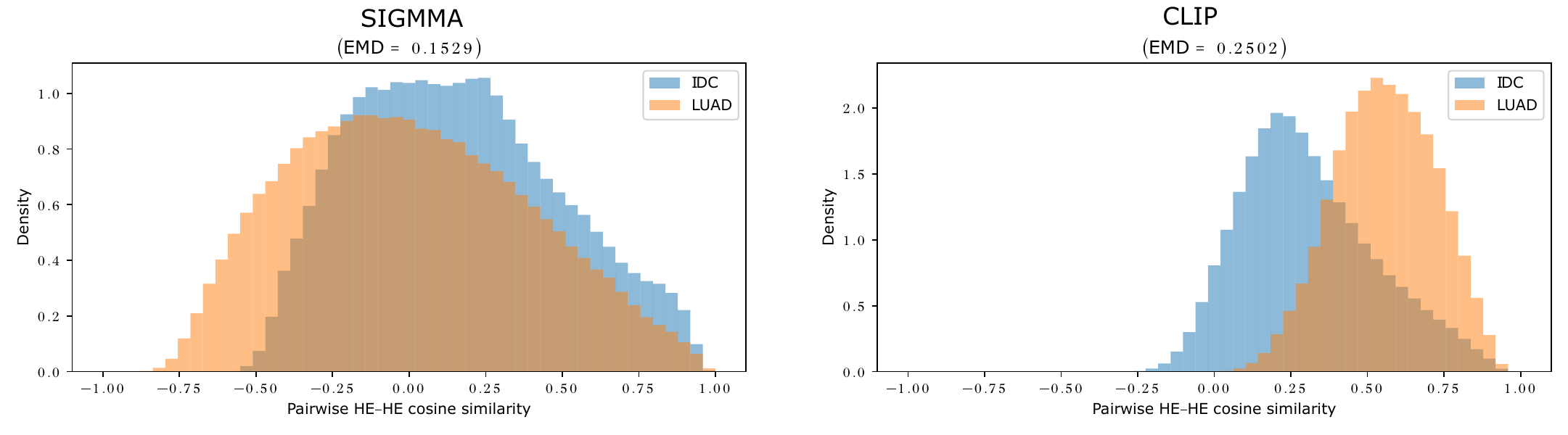}
  \caption{Cosine similarity of HE embeddings.} 
  \vspace{-10pt}
  \label{fig:HE-HE_similarity}
\end{figure}

\paragraph{Embedding similarity challenges IDC retrieval.}
As shown in Fig. \ref{fig:HE-HE_similarity}, HE embeddings in IDC from \textsc{sigmma} are highly homogeneous, making many tiles nearly indistinguishable and inherently limiting ST$\rightarrow$HE retrieval. By contrast, CLIP produces more dispersed HE embeddings for IDC, indicating greater apparent variability. This difference in feature distribution explains why ST$\rightarrow$HE retrieval drops for \textsc{sigmma} specifically on IDC. 

\subsection{Limitations}
Although \textsc{sigmma} learns meaningful multi-modal representation, several limitations remain. The model's generalizability is constrained by the limited range and diversity of available paired HE–Xenium ST datasets, hindering robust performance across heterogeneous tissue types. Additionally, because the approach relies on hierarchical spatial graphs constructed from cell segmentation, its effectiveness is inherently dependent on segmentation quality, which may be variable in complex tissue contexts. 

\subsection{Future work}
Future extensions of \textsc{sigmma} include evaluating the framework on other single-cell–resolution spatial transcriptomics platforms (e.g., CosMx, MERSCOPE) to assess cross-platform robustness. In addition, \textsc{sigmma}’s hierarchical design naturally scales to the WSI-level task, which we plan to explore as a next step. Finally, extending retrieval evaluation beyond within-tissue settings to cross-tissue scenarios may reveal how well the learned representations generalize across distinct morphological and molecular contexts.



\end{document}